# Agent-based Modeling and Simulation of Human Muscle For Development of Human Gait Analyzer Application

Sina Saadati[1], Mohammadreza Razzazi[2]

## Abstract

Despite the fact that only a small portion of muscles are affected in motion disease and disorders, medical therapies do not distinguish between healthy and unhealthy muscles. In this paper, a method is devised in order to calculate the neural stimuli of the lower body during gait cycle and check if any group of muscles are not acting properly. For this reason, an agent-based model of human muscle is proposed. The agent is able to convert neural stimuli to force generated by the muscle and vice versa. It can be used in many researches including medical education and research and prosthesis development. Then, Boots algorithm is designed based on a biomechanical model of human lower body to do a reverse dynamics of human motion by computing the forces generated by each muscle group. Using the agent-driven model of human muscle and boots algorithm, a user-friendly application is developed which can calculate the number of neural stimuli received by each muscle during gait cycle. The application can be used by clinical experts to distinguish between healthy and unhealthy muscles.

**Keywords:** Agent-based Modeling and Simulation; Artificial Intelligence in Medicine; Human Muscle Modeling; Biomechanical Intelligent Application.

## 1. Introduction

Muscle is a key organ of human motion. It receives neural stimuli known as Action Potentials (ACs) and generates mechanical force as reaction. In many motion disorders, including Parkinson's Disease or MS, this process can be violated. In these cases, the neural system of patient is defected and, as a consequence, human experiences some problems in his muscle's contraction. Motion disorders can be detected by physicians and clinical experts using simple or advanced methods. In some cases, patients are asked to move their bodies or walk while in some other cases, Electromyography (EMG), which is a painful tool, is used. EMG is an equipment which enable experts to estimate the intensity of neural signals that are being received by muscles. Surface EMG (sEMG) is another tool that attempts to capture EMG from the skin. Despite EMG, electrodes are not entered inside the body in sEMG. While sEMG is not painful, it is not accurate because noise signals are grown if the distance between muscle and electrode is increased. After detection of the disease and its intensity, some medical treatments are provided to the patients. Despite the fact that in motion disorders, only a portion of muscles are affected, medical therapies influence all the muscles even muscles of the eyes. As medical treatments include chemical drugs which have harmful consequences, It is important that clinical experts distinguish between healthy and unhealthy muscles. It can be done using a user-friendly application which is based on accurate and scientific methods.

Computerizing medical processes is a method causing an increase in the accuracy of medical and clinical processes and a decrease in the costs [1,28,39]. Haleem et al. reviewed researches concentrating on computerizing medical processes including detection, treatment, and even operation [8]. Medical image analysis is a popular field in which researchers are developing convolutional neural networks to detect and localize the cancer pathologies including breast cancer and so on [43]. Prediction and detection of heart disease can be done using computerized methods [42]. In some Artificial Intelligence (AI) methods, especially deep learning, the software is not interpretable causing the calculations ambiguous [5,41]. Nezhat et al. have anticipated that by the

[1] Department of Computer Engineering, Amirkabir University of Technology, Tehran, Iran, Sina.Saadati@aut.ac.ir
[2] Department of Computer Engineering, Amirkabir University of Technology, Tehran, Iran, razzazi@aut.ac.ir

2050 more than 90% of medical operations will be done by computers [33]. While having powerful methods in order to predict, detect, and treat medical disorders, development of a user-friendly application must be considered because program is not known fruitful if no one can use its facilities [4,31,38]. Also time-complexity and computational efficiency must be considered to minimize the energy-consumption of the model, especially in the cases like intelligent prosthesis that battery is used [40].

Gait cycle has a key role in detection of the kind and intensity of motion disorders including Parkinson's Disease or MS [37]. Regarding the subjective inherent of clinical classification and rating, many researches have been done to computerize a part of this aim in order to lower the human error [30,34]. Lempereur et al. have devised a method that can distinguish different phases of gait cycle for a child [23]. Zhang et al. have proposed a method which enable us to estimate ankle joint torque using EMG signals using neural networks [50]. But, it must be considered that neural networks are so-called black box and are ambiguous. Also, EMG signals are not exact neural stimuli that cause muscles to contract. EMG signals are only an estimation of neural activity in the muscle site. In addition, neural network models can investigate the motion in a limited situation while the proposed model of human muscle in this paper can be surveyed in any possible situation. In [44] strain gauge sensors are used in order to measure leg muscles activity during gait cycle, which is not as accurate as other methods like those that use markers.

In this paper, a method is proposed aiming at calculation of neural stimuli entered to muscles of lower body in a gait cycle. For this purpose, an agent-based model of human muscle is devised to simulate behavior of any type of human muscles. It can compute generated force caused by muscle contraction based on its received ACs. The structure of the model is designed as the process can be done vice versa. Since the model has effective computations, it can also be used in intelligent prosthesis where battery-friendly is under consideration. Then, Boots Algorithm is proposed in order to calculate forces that have affected the skeleton based on angles of lower body joints during gait cycle. Finally, using Boots Algorithm and agent-driven model of human muscle, it is possible to compute neural stimuli entered into each muscle by having lower body joint angles in a gait cycle. Therefore a user-friendly infrastructure-independent application is designed and developed to facilitate use of the method for clinical experts. It is recommended that physicians and experts distinguish between healthy and unhealthy muscles using the proposed application. During this research, some libraries have been developed which are published as open-source programs that enable modelers and researchers to demonstrate a 3D graphical model of human with only calling three functions, making it easy-to-learn.

In the second section, preliminaries are explained including definition and principles of agent-based modeling and simulation in order to design the roadmap for development of the model, physiological structure of human muscle to emphasize that the proposed model is based on the exact anatomy of muscle in spite of neural network models which are not interpretable, and biomechanical principles that fill the gap between muscle contraction and the movement. In the third section, the agent-based model of human muscle is described. In addition, Boots algorithm is explained which functions as a biomechanical model.

## 2. Preliminaries

In this section, preliminaries are explained. For designing and developing an agent-based model of human muscle, it is necessary that readers be aware of agent-driven modeling and simulation, physiological structure of human muscle, and biomechanical equations. In section 2.1 definition of agent-based modeling is described. Differences between agent-driven modeling and other types are explained and the reason why this kind of simulation is suitable for human muscle is emphasized. In section 2.2 the physiological structure of human muscle is described to emphasize that the proposed model is based on what is being done in the nature. In 2.3 biomechanical equations are explained in order to be used in Boots algorithm and at last, auser-friendly application is developed in order to analyze human walking. Finally in the fourth section, discussion and future works are described.

### 2.1. Agent-based Modeling and Simulation (ABMS)
Modeling and simulation is a process that researchers aim at reconstructing a phenomenon in software. This process consists of two phases in which researcher attempts to describe the behavior of the phenomenon using mathematics which is named modeling and then, he develops a software that can recreate the case in a computer program which is known as simulation. Agent-based modeling and simulation concentrate on the autonomy of the elements called agents in spite of other types of simulation that each element can be managed by an outer element like process-based or equation-based modeling and simulation [17,26]. Agents are located in an environment and can gather some information from their surrounding location or interact with other agents. One advantage of ABMS is the philosophy behind the process which highlights the fact that in the nature, all the elements including humans, animals, objects, and even stars are autonomous and behave independently [18,21,24,47]. Therefore, by noticing this point, our simulation would be natural meaning that the algorithm executing inside the computer is similar to the algorithm which is being done in the real world, and, as a consequence, results would be more accurate and reliable. Hence it is recommended that researchers use agent-driven methods for simulating medical and biological studies [32].

Simulation of an agent-based model can be done using prefabricated tools including Repast Simphony or AnyLogic or it can be developed using a programming language. Since ABMS is equivalent to Object-oriented programming, it is suggested that modelers use object-oriented programming languages to simulate agents. As there is no need to learn a new modeling language and gain skills in using the new tool, this method would not be time-consuming [3,27,32,47]. Java is a well-known object-oriented programming language which is also infrastructure-independent meaning that programs built by Java are able to be executed on any type of computer regardless of its hardware architecture or its operating system. Additionally, many ABMS languages are based on Java including Repast Symphony or SPARK [45]. Consequently, Java is chosen as the base for developing the proposed agent-based model of human muscle.

## 2.2. Physiological structure of human muscle

Motion is human and animals pursue a process in which neural signals known as ACs are sent from the brain or spinal cord to the muscles stimulating them to be contracted. This process can be done unconsciously which destination muscles are known as involuntary muscles or consciously that muscles are named voluntary muscles. In this paper, voluntary muscles are considered although the proposed method can be followed to simulate involuntary muscles in some cases. The physiological structure of human voluntary muscle is demonstrated in Figure 1. Muscles are made up of many string-like elements known as fascicles that are connected to each other by a protein called perimysium. Each fascicle contains a number of muscle fibers covered by a shield called sarcolemma. They are linked together by a connective tissue called the endomysium. Each muscle fiber is biologically a cell. It contains many tiny strings called myofibrils. Each myofibril contains many paired elements called actins and myosins. The nervous stimulating signals or ACs cause calcium ions to be released inside myofibrils, which causes chemical interactions between actins and myosins, and as a result, myofibrils and muscle fibers contract until ions are completely depleted [6,9,25,49]. Contraction causes a physical force affecting the bones and skeleton. Finally, following the biomechanical laws, movement happens. This complex is known as the neuromusculoskeletal system.

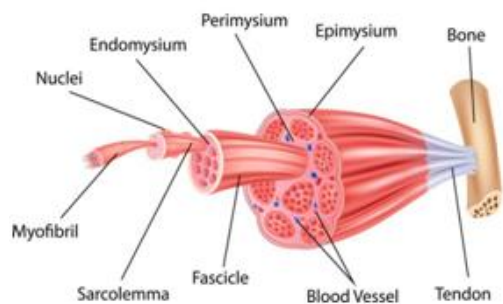

Figure 1. The physiological structure of human voluntary muscle [16].

In the neuromusculoskeletal system, one neuron is connected to multiple muscle fibers forming a Motor Unit (MU). One AC from the neuron is sent to all the fibers in that MU and all of fibers, not a portion of them, will be contracted as the reaction to the AC. Neural system can manage the strength of the contraction by stimulating more MUs. Also, in one MU, contraction is empowered if the frequency of the ACs sent by the neuron is increased, because each time AC is conveyed to muscle fibers, calcium ions are released. If subsequent ACs are transmitted to the fibers before the depletion of calcium ions, there will be more ions inside the myofibrils and, hence, the contraction is intensified. The process can be repeatedly done until all the actins and myosins inside the MU are stimulated by calcium ions. At this time, MU is producing its maximum contraction force and surge in the ACs frequency will not cause more force generation. MUs can be classified in a spectrum in which one type of MUs known as fast-twitch react to AC very quickly in spite of the slow-twitch kind that has moderate reaction. Fast-twitch MUs are relaxed instantly after AC while slow-twitch MUs have gradual relaxation. According to Henneman's Size principle, in a muscle contraction, fast-twitch MUs are stimulated before slow-twitch MUs to minimize the energy consumption of the muscle [7,10,11,13]. The structure of muscle and process of contraction which is described is basis of the proposed agent-driven model of human muscle making it natural and reliable. In addition, difference between ACs and EMG is considered meaning that EMG is an estimation of all ACs in a position which there is also the risk of noise inclusion.

### 2.3. Biomechanics Engineering

The force generated by the muscles does not cause similar movements, because the skeleton and bones behave as mechanical levers. Biomechanics is the study of the movement of humans and other fauna in which the physical and mechanical rules over the body are cogitated. It is made up of two fields: kinetic and kinematic. Kinetics investigates the reasons for the movement, whereas kinematics considers the quality of the motion. In the analysis of human motion, several important formulas are used. In equations 1 to 4, τ, F, d, θ, I, α, v, and t signify torque, force, the distance between force and rotation center, the position of the rotating object, the moment of inertia, angular acceleration, angular velocity, and time, respectively [6,19,25,49].

$$\tau = F.d \quad (1)$$

$$\tau = I\alpha \quad (2)$$

$$\alpha = \frac{\Delta v}{\Delta t} \quad (3)$$

$$v = \frac{\Delta \theta}{\Delta t} \quad (4)$$

The human gait cycle is an essential topic in biomechanics. It consists of some actions followed by a human during walking. Generally, a gait cycle has two phases: the stance phase, in which a leg is located on the ground and supports the human's weight, and the swing phase, that the leg moves and is detached from the ground. The gait cycle is a key for physicians to diagnose motion disorders like Parkinson's disease and MS and clarify intensity of the illness [2,15,17,46].

## 3. Proposed Method

In this section, a method which make us able to calculate the neural stimuli from the angles of lower body joints using a reverse engineering is explained. Agent-based model of human muscle is proposed in order to calculate the neural stimuli or ACs that have been entered into the muscle from the forces it has generated. Boots algorithm is then described to compute the forces which are affected the bones from the lower body joints angles during a gait cycle. Figure 2 illustrates the process of movement in a human compared with the proposed method for reverse engineering of the process.

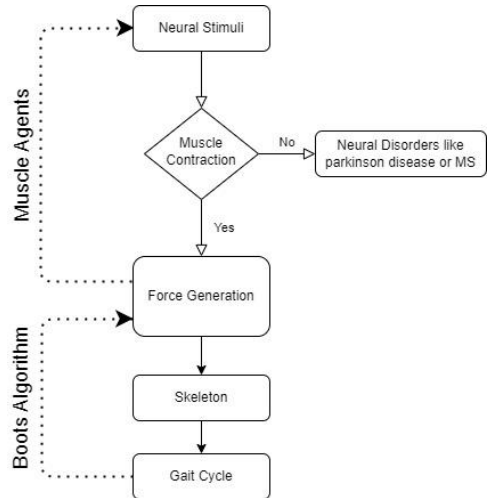
Figure 2. Flowchart of human motion and the proposed method.

## 3.1. Agent-based modeling and simulation of human muscle

In the proposed method, each muscle was viewed as an agent that could be contracted independently. Figure 3 depicts the structure of the agent-based model of human muscle. The model contains two independent internal agents, including active and passive muscle parts. The active part, composed of MUs agents, acts in response to nervous stimuli (ACs), whereas the passive part generates contraction force based on the elasticity feature of the muscle [35,36]. The passive agent's behavior is expressed by Eq.5, where $F_{P0}$ is a constant value depending on the type of the muscle and l is the muscle's length compared to the situation when it is in the relaxed state [20,22,49].

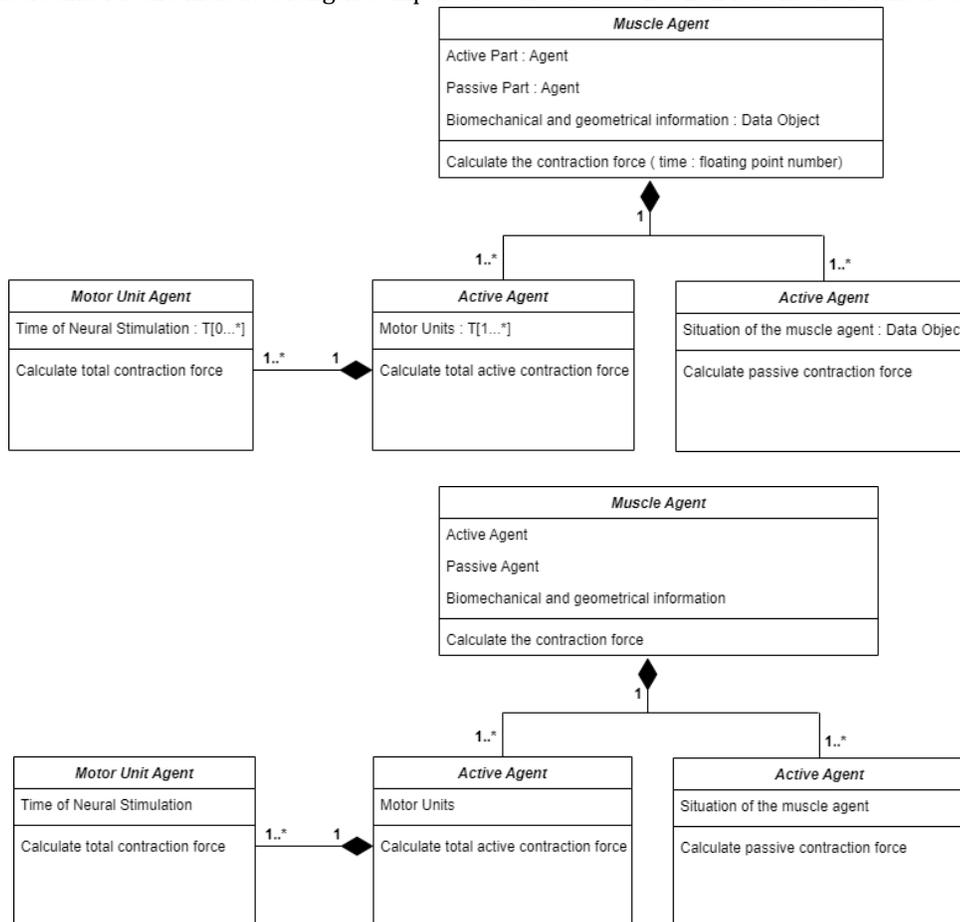
Figure 3. The structure of the human muscle model.

Receiving an AC causes the active agent, which is controlled by the neural system, to contract. After the AC is transferred into the muscle fibers, calcium ions are released inside the myofibrils, causing a chemical interaction between actins and myosins, resulting in a contraction force. Recruiting more MUs or increasing the neural stimulation frequency will generate more force. [48] proposed a formula for explaining the force generation of a single MU after a single stimulus. The formula is shown in Eq.6, where $F_0$ is a constant value that varies depending on the muscle's kind, T is the time between the stimulation and MU tension peak which is based on the type of the MU, and t is the current time. In this equation, T is supposed to be higher in stronger muscles like hamstrings. It can also be affected by factors such as fatigue or ambient temperature.

$$F_{Passive}(l) = Maximum\{F_{P0}\, e^{(l-1)} - 1\, , 0\} \quad (5)$$

$$F_{winter}(t) = F_0 \frac{t}{T} e^{(\frac{-t}{T})} \quad (6)$$

In Eq.6, the duration of contraction is not considered. The MU contraction can be continued from 10 milliseconds to 100 milliseconds [6,9,10,29]. In contrast, according to Eq.6, a typical MU in the lower body muscles can keep contracting for more than 600 milliseconds by a single stimulus which does not correspond to the valid information. Therefore, in this study, we proposed a new equation for analyzing MU behavior after receiving a single stimulus, as shown in Eq.7, where $F_0$, T, and t are defined as in Eq.6.

$$F_{sina}(t) = F_0 \frac{t}{T} t^{(\frac{-t}{T})} \quad (7)$$

It is known that increasing the contraction duration of a MU reduces the maximum peak of its tension force. This phenomenon can be seen in Eq.7 by changing the value of $T$ with a constant amount for $F_0$. Eq.6 and Eq.7 describe the tension force of a MU after receiving a single AC. As mentioned in the section 2.2, if the muscle is stimulated again before it is relaxed (when the generated force due to the Eq.7 is negligible.), the process of releasing calcium ions in the myofibrils will be repeated. As the calcium ions released by the first stimulus are not obliterated yet, more calcium ions will be inside a myofibril, causing more interaction between actins and myosins and, consequently, more intense contraction. This phenomenon, known as wave summation, is expressed by Eq.8, where $t$ represents the current time. An increase in stimulation frequency will only sometimes increase tension force. Spreading more calcium ions will be ineffective if all actins and myosins within a MU are interacting. This situation, known as tetanus, is described by Eq.9, where $MU\_Maximum\_Force$ is the maximum tension force that a MU of its type can generate [6,10,11]. The behavior of MotorUnit agents can be defined and simulated using the Eq.9.

$$F_{multiStimuli}(t) = \sum_{for\ all\ stimuli\ occured\ in\ t_j} F_{sina}(t - t_j) \quad (8)$$

$$F_{MU}(t) = Minimum\{F_{multiStimuli}(t)\, , MU\_Maximum\_Force\} \quad (9)$$

Another factor causing a muscle to generate stronger tension is the number of MUs recruited by the neural system. It is also important that a muscle's length significantly influences the tension force. Figure 4 depicts the relationship between length of the muscle and active tension force, which is mentioned as $R_{length-force}(l)$ function in this paper. As a result, Eq.10 can describe the behavior of a muscle's active agent. Finally, Eq.11 explains the operation of a muscle agent. In both equations 10 and 11, $t$ denotes the current time, and $l$ refers to the length of the muscle in the current situation compared to its length in the relaxed state [6,9,14,19,49].

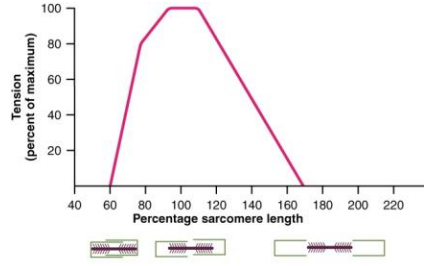

Figure 4. length-tension relation of a muscle [14].

$$F_{active}(t,l) = R_{length-force}(l) \times \sum_{for\ all\ MU\ agents} F_{MU}(t) \quad (10)$$

$$F_{muscle}(l,t) = F_{Active}(l,t) + F_{Passive}(l) \quad (11)$$

The agent-driven model was inferred using deductive reasoning methods, in which the researcher discovers new facts based on previously accepted laws and facts. Therefore, the simulation results based on the proposed model are expected to be correct and reliable. In addition, the model is simulated to investigate its behavior. Since a biomechanical model is necessary for this simulation, it is assumed that each joint of human's lower body is connected to a pair of muscle groups as flexor and extensor. In total, six independent groups of muscles are supposed to influence three joints involving ankle, knee, and hip for each leg. As in this section it is aimed at checking the agent-driven model of human muscle, the assumption of independence between muscle groups is regarded only to simplify the calculations. In the section 3.2, this assumption is improved by providing a fine biomechanical model. The designed biomechanical model in this section is illustrated in Figure 5 in which muscles are depicted by green and red lines showing flexors and extensors, respectively. For ankle joint, as it can be understood in the figure 5, dorsiflexion is considered as flexion and plantarflexion is contemplated as extension. Subsequently, angular movements of lower body joints are used as the input and based on the biomechanical model, angular accelerations of each joint is computed by inverse dynamic methods. For the reason that the number of variables is more than number of equations in this case, optimization methods are used to make the problem solvable. Finally, using muscle agents, neural activities of each muscle group is calculated. In the other words, the steps of the process shown in Figure 6 are passed opposite. The calculations are done regarding that the movements out of the sagittal plane are negligible to simplify the computations.

As muscle activity has a direct relation with neural stimulations, neural activity of the groups are compared with the muscle activity diagrams measured by other researchers (as is demonstrated in Figure 12). A portion of muscles in the lower body affect two joints at the same time. For example, hamstrings are responsible for movements in the knee, hip, and pelvis. Therefore, the activity of these types of muscles is separated into different periods based on the goal the muscle is contracting for. For example, if hamstrings are contracting in a period like P1 to do an extension in the hip joint and they are also being contracted in a period like P2 to do a flexion in knee joint, their activity during P1 is compared with the neural activity of the hip extensor muscle group and their activity in P2 is compared with the neural activity of knee flexor muscle group. The comparison shows that the muscle activities calculated in this section are compatible with the information measured by other researchers because at each period, both comparing muscle groups are active (or inactive) with same intensity. Regarding the fact that neural stimulations have been assumed as the muscle activities, it can be claimed that the agent-driven model of the muscle is correct and reliable. Finally, the whole process is reversed meaning that an agent-based simulation is developed which receives calculated ACs (representing neural stimulations) as the input and displays graphical 3D human motion as the output as is shown in Figure 7. The output displays a person walking normally without any error or shaking. The process of the simulation is demonstrated in Figure 6. The simulation is developed using JavaScript programming language which makes the simulator able to be executed on any kind of computer including personal computers, intelligent mobiles, and even smart televisions without installing any necessary software or libraries. Therefore, the simulator is user-friendly and easy-to-use.

For displaying the 3D graphical model of human moving in a user-friendly environment, ThreeJS library is used in the simulation described in this section. The library is easy-to-learn and as it is based on JavaScript, it is infrastructure-independent, too. For this reason, the simulation is programmed by JavaScript programming language. As explained in the section 2.1, object-oriented programming languages like Java are recommended for programming an agent-driven simulation. Therefore, a Java library is designed and developed in order to display motions of a human in a 3D graphical environment using ThreeJS library. The library which is published as open-source program, does not need any prerequisite application or library and is as easy-to-learn and transparent as the programmer does not need to know anything about JavaScript language. Java programmer is able to use the library by calling only two or three simple functions. It is explained in the section 3.3 with more details.

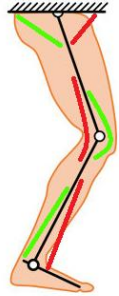

Figure 5. a simple biomechanical model of human lower body.

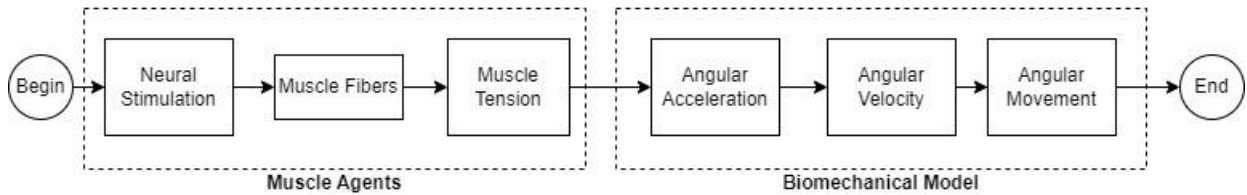

Figure 6. Strategy of the agent-based simulation of human gait cycle.

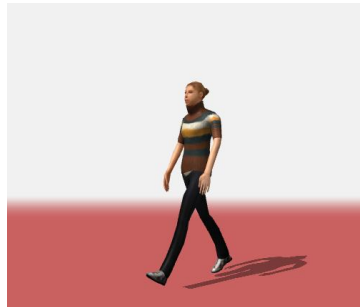

Figure 7. The output of the simulation.

The proposed agent-based model of human muscle is natural and effective. As the model's calculations follow O(1) time complexity, it is recommended that the model be used in the intelligent prostheses where battery consumption is critical. In this case, it is possible to assume that there is a direct relation between intensity of EMG (or sEMG) signals and the number of ACs. Then, generated force of muscle contraction can be predicted using the model. It must be emphasized that the model is also designed flexibly as any type of muscle like muscles of the leg or eyes, can be simulated using the proposed model because it is precisely based on the structure and behavior of voluntary muscles of human. For this aim, it is sufficient that variables like $F_0$ or $T$ in the Eq.7 be changed and configured. Even different types of MUs including fast-twitch or slow-twitch MUs or different situations like when the muscle is affected by a freezing environment can be simulated with the proposed model. For example, amount of the variable $T$ in the Eq.7 is increased if the temperature is decreased [49].

### 3.2. Boots Algorithm

Generally, researchers used two methods to calculate muscle activity: electromyography, which is painful, and reverse engineering (also known as inverse dynamics in biomechanics), which is based on mathematical optimization solutions that are inaccurate in many studies as the output of the equation is attributed to one variable (or a limited set of variables). In other words, the challenge of optimization methods is that in a typical joint, only one muscle can be considered active, which is only sometimes the case.

In this section, a solution to this challenge is proposed. Boots algorithm allows researchers to calculate the activity of lower body muscles during the gait cycle using reverse engineering by decreasing the limitations of optimization solutions. The Boots algorithm takes as input the angular displacement of three joints: hip, knee, and ankle in the sagittal plane. The algorithm's output is the tension force produced by each muscle in the human lower body. The algorithm is planned using an accurate human biomechanical model and accepted physiological facts [6,9,25,49]. Figure 8 provides a general view of the components of the algorithm. As can be seen, the lower body is assumed to have 6 muscle groups including triceps surae, tibialis anterior, hamstrings, quadriceps, gluteus, and iliopsoas muscles. While most groups are only responsible for the movements occurring in one joint, hamstrings and quadriceps have the responsibility of moving knee, hip, and also pelvis [6,9,25,49]. Regarding that the upper body mass center is swing compared to the location of pelvis, hamstrings and quadriceps have to contract to keep the status of the pelvis. Boots algorithm is designed based on the philosophy behind each contraction of the muscles. It must be noticed that the neural system of the body controls the gait cycle intelligently to minimize the energy consumption [29]. In this scenario, the fact that two agonist and antagonist muscles can be contracted at the same is justified by the reason that they have different tasks simultaneously. With regard to the goals behind each contraction, limitations of optimization method can be lowered. As a consequence, the results would be more reliable and more similar to what is being happening in the real world. Due to this idea, boots algorithm is separated into different components, each one focusing on a single goal. For example, as is shown in Figure 8, the first phase of Boots_Hamstrings algorithm concentrates on the hamstring muscles that are responsible for the motion in the knee joint while its second phase focuses on the events relating to the hip joint that can be movements of thigh or pelvis.

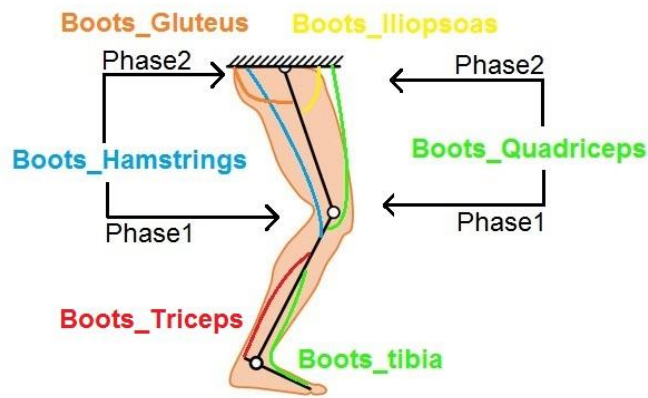

Figure 8. General structure of Boots algorithm.

Boots_Triceps algorithm calculates the force generated by triceps surae muscle in a gait cycle. This is the first component of the boots algorithm in which the design is based on the fact that bulkiness of a muscle is because of its need to generate stronger force. As triceps are responsible for negating the ground force reaction, they are huger compared to anterior muscles of the leg, which can be seen in the structure of boots. In addition, if you notice your own muscles, you will feel that the contraction of posterior muscles is much more than anterior tibialis muscle. The biomechanical model behind the algorithm is demonstrated in Figure 8. In the boots algorithm, the acceleration can be computed using the angular displacement by Eq.3 and Eq.4 and then, using Eq.1 the torque is computable.

Boots_Triceps Algorithm
Input: Angular displacement of ankle joint
Output: Tension force produced by Triceps Surae muscle
1   Create a vector with zero values named Results ;
2   TotalTorque = (ankle -> moment_of_inertia * ankle -> angularAccelerationDuringGaitCycle) ;
3   GroundForceReaction = Compute the ground force reaction using the body mass and position of the leg during the gait cycle.
4   **For** each step in the gait cycle like s do{
5     Moment_ground = based on GroundForceReaction[s] and considering that the ground force reaction is applied to the feet toe and using Eq.1, compute the moment generated by ground force reaction.
6     Moment_feet = Considering the mass of feet regardless of thigh and shank, the moment produced by the influence of gravity on the feet is calculable.
7     Moment_muscle = TotalTorque[s] - (Moment_feet + Moment_ground ) ;
8     Results[s] = Based on Eq.1 and the value of Moment_muscle, It is possible to compute the force produced by the muscle.
    }
9   **Return** Results ;

Boots_Tibia algorithm calculates the force that is generated by tibialis anterior muscles in a gait cycle. Since tibialis anterior can only do dorsiflexion in the ankle joint, it mostly controls status of the ankle which can be investigated in two situations. The first situation is at the beginning of the stance phase when the heel is located on the ground that tibialis muscles attempt to prevent toes from hitting the ground. In some patients suffering from MS or Parkinson's disease, this muscle is not strong as toes hit the ground very hard. The second situation is at the swing phase that the mass of the foot causes the ankle to do a plantarflexion. In this circumstance, tibialis muscles contract to keep the stability of the ankle. Described states are investigated in the first and second phases of the Boots_Tibia algorithm, respectively.

Boots_Tibia  algorithm
Input: Angular displacement of ankle joint
Output: Tension force produced by Dorsiflexor muscle
1     Phase1 = Boots_Tibia_phase1( );
2     Phase2 = Boots_Tibia_phase2( );
3     **Return** VectorSum( Phase1 , Phase2 ) ;

Boots_Tibia phase 1 algorithm
Input: Angular displacement of ankle joint
Output: Tension force produced by Dorsiflexor muscle
1     Create a vector with zero values named Results ;
2     TotalTorque = (ankle -> moment_of_inertia * ankle -> angularAccelerationDuringGaitCycle) ;
3     GroundForceReaction = Compute the ground force reaction using the body mass and position of the leg during the gait cycle.
4     **For** each step in the gait cycle, like s {
5       **If**( GroundForceReaction[s] < feet.weight ){
6           Results[s] = Compute the tension force of dorsiflexor muscle using the value of  Torque[s], weight force of the feet and Eq.1.
7       }**else**{
8           Results[s] = 0 ;
        }
      }
9     **Return** Results ;

Boots_Tibia phase 2 algorithm
Input: Angular displacement of ankle joint
Output: Tension force produced by Dorsiflexor muscle

1   Create a vector with zero values named Results ;
2   TotalTorque = (ankle -> moment_of_inertia * ankle -> angularAccelerationDuringGaitCycle) ;
3   GroundForceReaction = Compute the ground force reaction using the body mass and position of the leg during the gait cycle.
4   **For** each step in the gait cycle, like s {
5       F_gr = Compute vertical force applied from the ground to the heel by value of GroundForceReaction[s].
6       **If** toes are not located on the ground{
7           Results[s] = Compute the tension force of dorsiflexor muscle using the values of F_gr, feet.weight and Torque[s] ;
8       }**else**{
9           Results[s] = 0 ;
        }
    }
10  **Return** Results ;

Boots_Quadriceps and Boots_Hamstrings are two other components of the boots algorithm focusing on the similar muscles in the lower body which have same biomechanical tasks, but in opposite directions. Due to the Kinesiological facts, they are not only responsible for movements of the knee and hip joints, but also controller of the stability of the pelvis during gait cycle. Considering this point enables us to investigate the simultaneous contraction of the lower body muscles during the gait cycle. Boots_Quadriceps algorithm is divided into two components focusing on the knee joint and hip joint, respectively. In the first phase of the algorithm, quadriceps attempt to extend the knee or keel it extended and straight. This is important in the middle of the swing phase of the gait cycle when the person intends to put his leg on the ground and also, during the stance phase that knee joint is holding the weight of the body by negating the ground reaction force. Both situations are considered in the first phase of the algorithm. The second phase of the algorithm contemplates the hip joint and pelvis. As mentioned before, Quadriceps aim at holding the pelvis stable. Therefore, the mass center of the whole body except the leg which is being investigated must be calculated. For this reason, position of the other leg that is in the swing phase, hands, etc must be regarded to compute their mass centers. Then, based on the horizontal distances of computed mass centers, the torque generated by quadriceps can be calculated as the muscles contract to negate the gravity and make the body and pelvis stable.

Boots_Quadriceps Algorithm
Input: Angular displacements of the knee and hip
Output: Tension force generated by Quadriceps muscle

1   Phase1 = Boots_Quadriceps_Phase1( ) ;
2   Phase2 = Boots_Quadriceps_Phase2( ) ;
3   **Return** VectorSum(Phase1 , Phase2) ;

Boots_Quadriceps Phase1 Algorithm
Input: Angular displacements of the knee
Output: Tension force generated by Quadriceps muscle

1   Create a vector with zero values named Results ;
2   TotalTorque = (knee -> moment_of_inertia * knee ->angularAccelerationDuringGaitCycle);
3   GroundReactionForce = Compute the ground force reaction using the body mass and position of the leg during gait cycle.
4   **For** each step in the gait cycle, like s {
5       M_gr = According to the values of GroundReactionForce[s] and Eq.1, compute the torque generated by the ground.
6       M_w = According to Eq.1 and the weight of the shank, compute the torque produced by the mass of the shank.
7     temp = TotalTorque[s] - ( M_gr + M_w ) ;
8     Results[s] = compute the tension force produced by quadriceps muscle using the value of temp variable.
   }
9  **Return** Results ;

Boots_Quadriceps Phase 2 Algorithm
Input: Angular displacements of the hip
Output: Tension force generated by Quadriceps muscle
1     Create a vector with zero values named Results ;
2     GroundReactionForce = compute the vertical force applied from the ground during the gait cycle.
3     **For** each step of the gait cycle, like s {
4        Compute the position of the mass centre of the upper body related to the hip joint. It is noticeable that the mass centre of upper body is located in the posterior of the hip joint while person is in the anatomical state. Therefore, the mass center can be moved in the posterior or anterior of the pelvis in sagittal plane.
5        Compute the position of mass centre of another leg which is stated in the swing phase.
6        Compute the position of the whole upper body based on the information gathered in steps 4 and 5. This can be done using a simple weighted averaging and considering the weight of each part and horizontal distances. The angle of the hip body in elderly people must be considered.
7        M_w = Compute the torque value produced by upper body (and another leg that is in swing phase) using the information in step 6.
8        Quadriceps muscle can negate the torque calculated in step 7. So, It is possible to compute the torque generated by the quadriceps muscle. GroundForceReaction[s] must be also considered in this step.
9        Results[s] = Using the information from step 8 and Eq.1, It is possible to compute the tension force of the quadriceps muscle.
    }
10    **Return** Results ;

Since the hamstring muscles have similar tasks compared with the quadriceps, but in different directions, the proposed strategy for quadriceps muscle can be pursued to calculate the tension force generated by the hamstrings.

Since hamstrings and quadriceps are longer levers compared to gluteus maximus and iliopsoas muscles, they are considered as main controllers of the knee and hip joints. Gluteus maximus and iliopsoas have auxiliary responsibilities. This is important when a hip movement is required without any influence on the knee joint. With regard to this point, Boots_Gluteus algorithms is designed. The tension force generated by iliopsoas muscle can be calculated by following the same strategy as Boots_Gluteus algorithm with regard to the direction of vectors.

Boots_Gluteus Algorithm
Input: Angular Displacement in the hip joint
Output: Tension Force generated by gluteus maximus muscle
1     Create a vector with zero values named Results ;
2     TotalTorque = (knee -> moment_of_inertia * knee -> angularAccelerationDuringGaitCycle) ;
3     GroundReactionForce = Compute the ground force reaction using the body mass and position of the leg during the gait cycle.
4     Quadriceps = Using quadriceps algorithm, calculate the generated force by the quadriceps muscle.
5     Hamstring = Using the hamstrings algorithm, compute the tension force produced by the hamstring muscle.
6     **For** each step of the gait cycle like s do {
7        Compute the torque produced by the ground force reaction by GroundReactionForce[s] and Eq.1.
8        Using Eq.1 and the mass value of the leg, compute the torque value generated by the leg's weight.
9        Results[s] = Using the information calculated in steps 4,5,7 and 8, it is possible to calculate the torque value generated by Gluteus maximus muscle. Then, the tension force can be computed by Eq.1.

```
      }
10  Return Results ;
```

As gluteus maximus and iliopsoa muscles have similar duties, it is possible to calculate the produced force by following the steps of Boots_Gluteus algorithm.

Using the Boots algorithm, it is possible to reverse the process of movement and calculate the tension force produced by important muscles in the lower body. It is possible to develop software that analyzes the human gait cycle based on the proposed agent-based model of human muscle and the boots algorithm, which other researchers and physicians can use. The variety of muscles in the lower body is wider than what is considered in the boots algorithm. But this is not a problem violating the proposed method. The challenge for inverse dynamics of human gait cycle is only related to the muscles and their connected points and this fact causes the number of variables to be increased in the equation system. Thus, two or more muscles with the same connection points can be assumed as one variable in the system of equations. Boots algorithm attempts to solve the problem by minimizing the assumptions of optimization methods by focusing on the intelligence management of motor neural system and the philosophy behind each contraction during gait cycle. The results for muscle groups with certain endings can be divided between muscles based on their bulkiness.

### 3.3. Design and development of a gait cycle analyzer application
Agent-driven model of human muscle and boots algorithm pave the way to develop a software which can analyze the human motion in a gait cycle. Developing the application is challenging not only scientifically, but also technically. For this goal, some libraries are developed in order to solve problems relating to the programming. Then, based on the proposed strategy, an application is developed. Finally, a user-friendly interface is developed and added to the program to make it easy-to-use for any type of user including researchers and clinical experts.

### 3.3.1. Designing and developing a software library to facilitate the modeling processes
User-friendliness is one of the essential characteristics of good software. Software must include a graphical user interface allowing any type of user to interact with the software and its features. In the simulation software, the results must be understandable. As an object-oriented language, Java is recognized as an appropriate choice for developing agents and a simulator. However, there is no easy-to-learn and easy-to-use Java library for displaying a 3D human model.

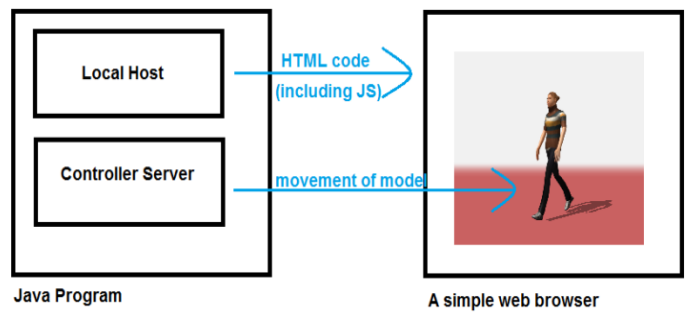

Figure 9. Architecture of the proposed library.

In biomechanics and human movement research, a graphical model of humans can make the results more understandable and realizable, so the software has been developed to be a user-friendly application. In this study, a Java library is built that allows the programmer and modeler to display a 3-dimensional graphical human with the ability to move her body. The library is composed of two web servers: a LocalHost and a ControllerServer. LocalHost occupies a determined web port and starts listening for any client request. It contains a code generator that builds a code in HTML that consists of JavaScript program and sends it to the client. The client can be browsers on the same machine or any browsers on other computers connected to the server via the network. The code sent to the client, which can be run on any web browser like Google Chrome without installing any other software, will launch an application that displays a 3D graphical model of a human. The graphical application will then connect to the ControllerServer and receive information about the model's last status, allowing the displayed model to be controlled by the Java library.

The library is published open-source, allowing other researchers and modelers to use it as a graphical infrastructure. The library is easy-to-learn because no new language is required to be learned. The modeler can use the library by calling only two functions thus its usage can facilitate and speed up the process of development of simulations. It is also easy to use because no additional libraries or applications must be installed. Figure 9 depicts the architecture of the library.

### 3.3.2. Design and development of gait cycle analyzer software

An agent-based modeling that receives neural stimulations and displays a walking person as an output described in section 3. The biomechanical model used in that modeling is based on simplified assumptions because the purpose in section 3 is to evaluate the agent-based model. Furthermore, because the reverse engineering method is based on mathematical optimization methods, the concurrent activity of muscles surrounding a single joint is not computable.

In this section, an application is developed to analyze the human gait cycle using the agent-based model and the boots algorithm. The application follows the flowchart of Figure 2 in a reverse order to calculate ACs entered into each muscle. First, the boots algorithm is used to investigate human motion during the gait cycle, allowing the angular velocity and angular acceleration of lower body joints to be calculated. Also, the tension forces produced by the lower body muscles are computed. The proposed agent-based model of human muscle is then used to measure the neural stimulation of each muscle. In this phase, Henneman's size principle is considered. Consequently, slow-twitch MUs are stimulated before fast-twitch MUs. In this study, a range of slow-fast twitch MUs is considered by defining four types of MU, as described in Table 1.

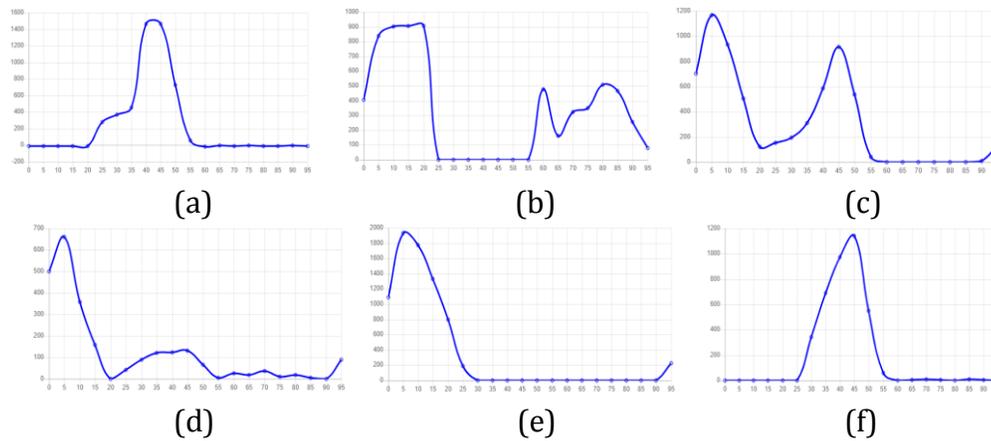

Figure 10. Activity intensity of lower body muscles during a healthy gait cycle. The triceps muscles. (a) Ankle's tibialis muscle.(b) Quadriceps muscle.(c) Hamstring muscle.(d) Gluteus maximus muscle.(e) Iliopsoas muscle.(f)

The application is used to compute the neural stimulation and corresponding activity of lower body muscles during a healthy person's gait cycle. Figure 10 shows the results, consistent with the valid information published by other researchers. Figure 10 provides graph diagrams with the horizontal axis representing the percentage of the gait cycle and the vertical axis signifying the relative tension force by muscles. In this study, the gait cycle begins when the heel is about to be displaced on the ground. Mentioning the starting point, it can be understood that outcomes are compatible with other published information.

Table 1. Classification of MU types.

| Class Name | Value of $F_0$ in Eq.7 | Value of $T$ in Eq.7 | About the class. |
|---|---|---|---|
| Type A | 0.1 | 20 | MU contracts very fast, but it is not durable. |
| Type B | 0.1 | 50 | MU contracts less fast and more durable than Type A. |
| Type C | 0.1 | 70 | MU contracts late, but durable. |
| Type D | 0.1 | 100 | MU contracts later than Type C and more durable. |

Figure 11 depicts lower body muscle activity during a healthy gait cycle based on previous measurements by other researchers using optimization methods. Figure 10 depicts the information obtained from this research that is compatible with the accepted information in Figure 11. Figure 10-a's graph diagram is compatible with the related diagram in Figure 11 as in both diagrams, there is a contraction period ranging from about 10% to about 55% of the gait cycle. The maximum peak in both diagrams occurs around 45% of the cycle.

The diagram shown in Figure 10-b is compatible with the diagram related to the tibialis anterior in Figure 11. It is noticeable that the tibialis anterior is an important dorsiflexor muscle in the ankle joint. Both diagrams show an activity period ranging from 55% of a cycle to about 25% in the subsequent cycle. The peak in both diagrams occurred between 0 to 20% of the cycle. It is also noteworthy that both diagrams show a trough between 55% and 100% of the cycle.

The diagram in Figure 10-c is compatible with its relative diagram in Figure 11 about the quadriceps muscle. Both diagrams have two maximum peaks. The first peak appears at the beginning of the cycle, between 90% in one cycle and 25% in the next. The second peak appears in the middle of the cycle. In the last stages of the cycle, there is a smooth increase in both diagrams. Notably, the influence of muscles such as the hamstrings or quadriceps on the pelvis is not investigated in the diagrams in Figure 11.

Lower body muscles activity is also surveyed using EMG measurements, as illustrated in Figure 12. Figures 12-e and 12-f show graph diagrams of hamstring muscle with a peak at the beginning of the cycle, consistent with Figure 10-d. In all Figures 12-e, 12-f, and 10-d, there is a weak but long-lasting activity between 20 and 60% of the cycle which is absent in Figure 11. As a consequence, the output information from the developed analyzer is more compatible with figure 12 which is based on EMG compared to information in Figure 11 that is based on optimization methods. This difference shows that the proposed method is successful in minimizing the error of optimization inverse dynamics [12]. All diagrams show an increase in the cycle's conclusion in all the diagrams mentioned. It should be noted that in this study, it is assumed that a gait cycle consists of 20 steps. Because the acceleration of each point is proportional to the angular displacement in three points, a 10% shift in the diagrams is to be expected.

The diagrams for the gluteus maximus and iliopsoas muscles in Figures 10 and 11 are compatible because they both have peaks in a similar span. It is noticeable that the results of the proposed method in Figures 10-a are compatible with the information provided by Figure 12-g which is related to the triceps surae muscle. The results are also consistent in Figures 10-b and 12-h, which are related to the ankle's tibialis muscle. The same is true for other diagrams relating to similar muscles between Figures 10 and 11.

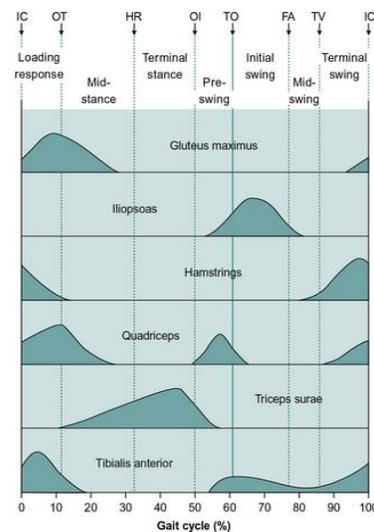

Figure 11. Relative activity diagram of lower body muscles during the normal gait cycle [46].

The proposed agent-based model can be used to analyze the operation of each muscle after the boots algorithm calculates tension force in the lower body muscles. In this step, a reverse engineering method is used in order to trigger different types of MUs listed in Table 1 based on the size principle to generate the tension force measured by the boots algorithm. It is possible to feed any angular displacement diagram of a lower joint, including the ankle, knee, and hip, to the developed application and investigate the stimulation of each type of MU in the lower body muscles as the software's output. Figure 13 depicts the stimulation of MUs for the ankle's dorsiflexor muscle during a healthy gait cycle. In the graph diagrams in Figure 13, the horizontal axis represents the time as a percentage of the cycle, and the vertical axis shows the stimulation intensity as number of ACs. Thus each vertical axis unit represents one action potential received by the MUs. It is noticeable that stimulation of type-A MUs in the tibialis muscle from 15% to 25% of the gait cycle is more than the same period for type-D MUs. This can be justified because the neural system requires intense and short-term tension, as the neural system can intelligently trigger MUs. It should be noted that investigation of MU types is impossible in electromyography studies.

### 3.3.3. Development of a user-friendly application for analysis of the human gait cycle

A good application is a result of a process that adheres to software engineering rules and principles. Having a user-friendly interface is one of the most critical requirements of any software. A powerful application is only helpful if the user can use its facilities. In this study, a user-friendly interactive graphical interface is designed.

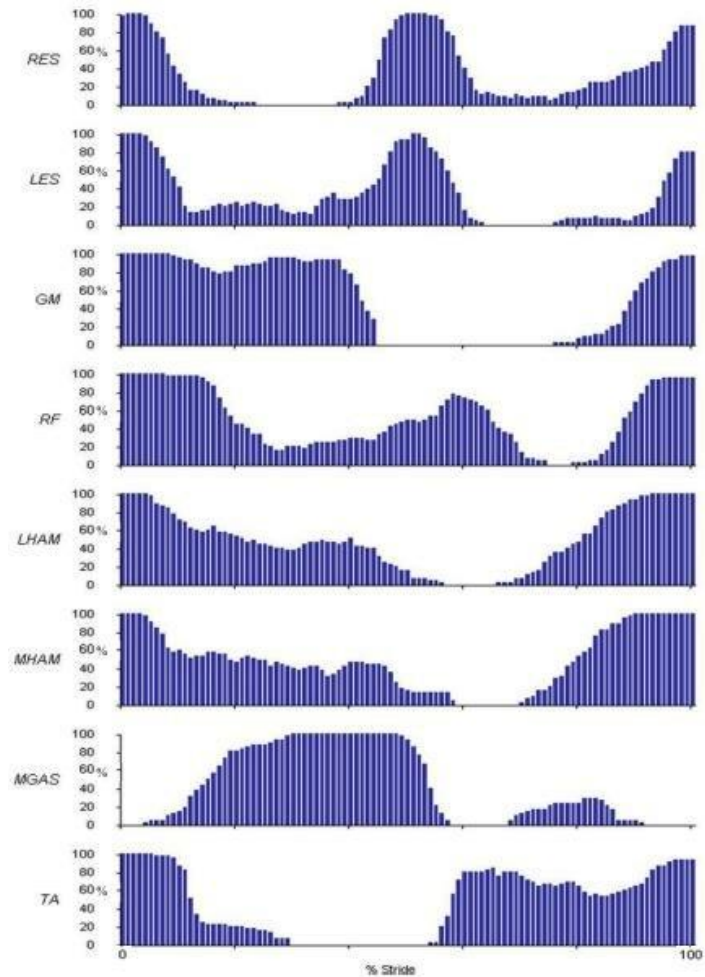

Figure 12. The electromyography activity of muscles during the gait cycle. The diagrams are related to the Right Erector Spinae muscle (RES), Left Erector Spinae muscle (LES), Gluteus Maximus muscle (GM), Rectus Femoris muscle (RF), Lateral Hamstring muscle (LHAM), Medial Hamstring muscle (MHAM), Medial gastrocnemius (MGAS) and Tibialis Anterior muscle (TA) [48].

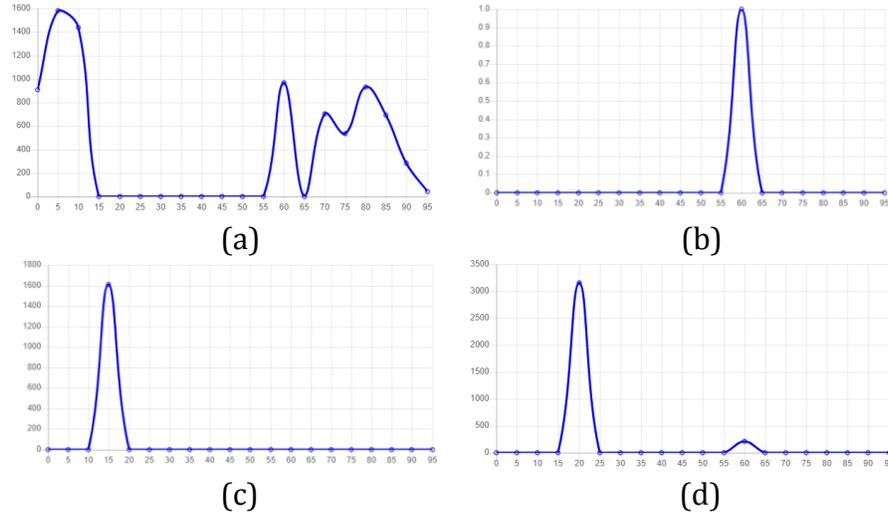

(a) (b) (c) (d)

Figure 13. stimulation of ankle's dorsiflexor muscle during a healthy gait cycle. The diagrams are related to MU with types A, B, C, and D, respectively.

The interface can display the motions during the gait cycle in a 3D space using the proposed library in section 3.3.1. Also, the interface can depict muscle tension force and MU stimulations using diagrams. Therefore, the results of the simulations will be understandable. Figure 14 demonstrates a part of the application that shows a walking person from various angles. In this program, the calculated ACs using reverse engineering are used as the inputs of lower body muscles simulated as agents. Then, each agent produces a mechanical force which is applied to the biomechanical model that is described in Figure 8 and human walking in a good-looking 3D graphical environment is the output of the software.

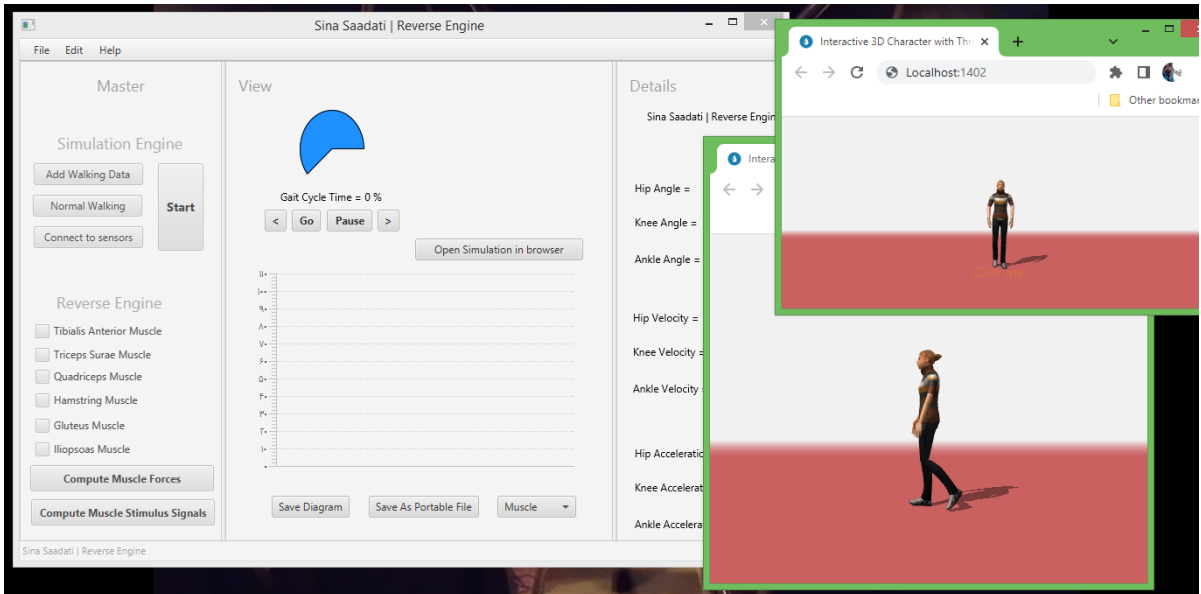

Figure 14. Graphical user interface of the application.

There is also the ability to simulate unhealthy gait cycles thanks to a powerful user interface. In this study, a walking disorder in which toes have an intense and abnormal collision with the ground at the start of the stance phase is simulated. This type of walking can be seen in many Parkinson's disease patients. Our findings indicate that the cause of this issue is the weakness of the tibialis muscles. In other words, in this type of walking disorder, only a limited set of muscles, such as the tibialis anterior muscle, are affected by the disease, while other muscles function normally. This phenomenon must be considered during treatment, particularly in some therapy methods that influence all muscles with chemical drugs. Figure 15 illustrates the comparison. Because many walking disorders are related to the neural system (like Parkinson's disease or MS), it is clear that in Figure 15, neural stimulation has been changed enormously while all the muscles have their natural stimulations. In the graph diagrams in Figure 13, the horizontal axis represents the time as a percentage of the cycle, and the vertical axis shows the stimulation intensity as number of ACs.

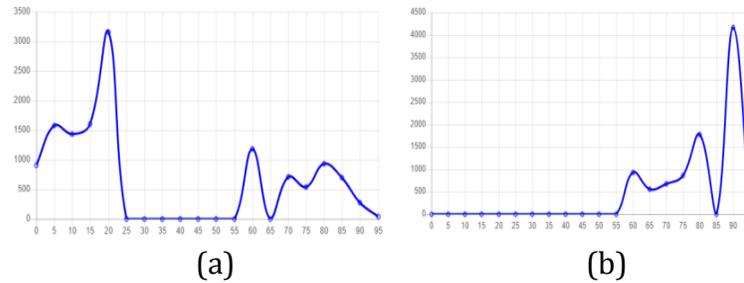

(a) (b)
Figure 15. Comparing stimulation of tibialis muscle during a healthy gait cycle(a) and unhealthy walking(b).

There are many popular applications with similar usage of the proposed method which might be mistaken with the developed program in this research. In Table 2, a comprehensive comparison has been explained. Using agent-driven method for modeling and simulation of human motion, which makes the calculations natural as is described in section 2.1, is one of the top priorities of this research. By having an agent-based methodology, we are able to calculate the number of stimulations for each independent muscle. Also, based on the software engineering principles, developing a user-friendly application that can be installed, learnt and used easily is another main priorities of the research in this paper. It is vital that software be easy-to-use because an application with great computational power, but has very complex interface which makes it impossible to be used by user is not valuable. During this research, authors experienced that there are many PhD researchers who claimed that good applications, including OpenSim, has as complex user-interface as it is very hard to be used. In the Table 2, UA stands for unavailable.

Table 2. Comparison of the developed software with other similar applications.

| Feature | OpenSim | AnyBody | Proposed Method |
|---|---|---|---|
| Infrastructure Independent | × | × | ✓ |
| User-friendly Interface | × | × | ✓ |
| Calculation of stimuli or EMG by reverse engineering method | × | × | ✓ |
| Distinguish stimuli based on MU type | × | × | ✓ |
| Calculation of both agonist and antagonist muscles of one joint in the lower body. | × | × | ✓ |
| Calculation Action Potentials entered to each muscle | × | × | ✓ |
| Distinguish Action Potentials based on MU types in one muscle. | × | × | ✓ |
| Ability of study for each kind of muscle(muscle of a child or elderly or a professional rider.) | ✓ | ✓ | ✓ |
| Ability of distinguishing healthy and unhealthy muscles in patients suffering from movement disorders. | × | × | ✓ |
| Using Artificial Intelligence and Machine Learning methods in order to analyze the movement. | UA | UA | In process |
| Ability of receiving data from markers connected to a patient or using motion signal processing. | ✓ | × | In process |

## 4. Conclusion and future works

In this paper, an agent-based model of human muscle was presented with a proposed boots algorithm which makes us able to calculate the forces produced by lower body muscles using a reverse engineering method. By comparing the activity diagrams of lower body muscles based on inverse dynamics that use optimization methods and EMG diagrams measured by other researchers with the results in this research, it has been shown that the proposed method covers the limitations of optimization inverse dynamics. Finally, a user-friendly application was designed and developed which gets the angular displacements of lower body joints including ankle, knee, and hip and calculates the neural signals stimulating the muscles. It can be used by clinical experts to distinguish between healthy and unhealthy muscles. The program is also able to simulate any kind of gait cycle with a good 3D graphical display. It must also be considered that despite neural network models, the proposed method is transparent as every component is clear and interpretable. In addition, using data can lower the flexibility of models in biomedical research because this can make the model dependent to the features of the people that have produces the data. These features include nationality, daily habits of the cases, age, gender and so on. So in the proposed method, design of the model is based on the structure of the muscle instead of data which include noise and include many hidden features to make the model flexible as it can simulate any kind of voluntary muscle. The described method in this paper solved a challenging problem. However, there are many options that can improve the application which will be studied in our future works. As walking is happening in a 2D sagittal plane, it is assumed that any force or physical variable that is out of the sagittal plane has a negligible influence on the process. However, the system will be more accurate if a 3-dimensional place is considered instead of a plane. Since artificial neural networks are not interpretable, in this paper it is suggested that experts use software with fully transparent components. It can be useful especially if the system needs to be improved. Although, neural networks can be used in many elements of the developed program in order to diagnoses detection, automatic drug-determination or intelligent advisor system who can recommend special sports movements in order to heal or improve unhealthy muscles. The input is assumed to be entered into the software using keyboard, however, it can be computerized by following a process in which intelligent mobiles are connected to the lower body organs and send their information from three-axis compass to the program using local network. Another idea is related to machine learning models that need data to be trained. It is suggested that data that are analyzed by the proposed software be stored in a cloud database to be used as train sets in developing machine learning models. Also, boots algorithm can be more accurate if the biomechanical model of human skeleton is defined more precisely. For example, the geometry of the bones can be considered in the model.

## 5. Declarations

### 5.1. Funding
Not applicable.

### 5.2. Conflicts of interest/Competing interests:
Not applicable.

### 5.3. Availability of data and material:
Not applicable.

### 5.4. Code Availability
You can access the developed models and sources at: https://github.com/sinasaadati95

### 5.5. Author's Contributions
The idea of generating an agent-based model of human muscle and using it to analyze human motion and distinguishing healthy and unhealthy muscles has been presented by the second author (Mohammadreza Razzazi). Modeling and simulation of human muscle and also, design and development of a simulator of human gait have been done by the first author (Sina Saadati). Boots algorithm is also presented by the first author.

# 6. References

[1] A. Bermeo, M. Bravo, M. Huerta, and A. Soto, "A system to monitor tremors in patients with Parkinson's disease," in *2016 38th annual international conference of the ieee engineering in medicine and biology society (embc)*, 2016: IEEE, pp. 5007-5010.
[2] E. Bonabeau, "Agent-based modeling: Methods and techniques for simulating human systems," *Proceedings of the national academy of sciences,* vol. 99, no. suppl_3, pp. 7280-7287, 2002.
[3] Ş. Bora and S. Emek, "Agent-based modeling and simulation of biological systems," *Modeling and computer simulation. London: IntechOpen,* pp. 29-44, 2019.
[4] M. Cardinot, C. O'Riordan, J. Griffith, and M. Perc, "Evoplex: A platform for agent-based modeling on networks," *SoftwareX,* vol. 9, pp. 199-204, 2019.
[5] I. Castiglioni *et al.*, "AI applications to medical images: From machine learning to deep learning," *Physica Medica,* vol. 83, pp. 9-24, 2021.
[6] R. T. Floyd and C. W. Thompson, *Manual of structural kinesiology*. McGraw-Hill New York, NY, 2009.
[7] T. Gordon, C. K. Thomas, J. B. Munson, and R. B. Stein, "The resilience of the size principle in the organization of motor unit properties in normal and reinnervated adult skeletal muscles," *Canadian journal of physiology and pharmacology,* vol. 82, no. 8-9, pp. 645-661, 2004.
[8] A. Haleem, M. Javaid, and I. H. Khan, "Current status and applications of Artificial Intelligence (AI) in medical field: An overview," *Current Medicine Research and Practice,* vol. 9, no. 6, pp. 231-237, 2019.
[9] J. E. Hall and M. E. Hall, *Guyton and Hall textbook of medical physiology e-Book*. Elsevier Health Sciences, 2020.
[10] E. Henneman, G. Somjen, and D. O. Carpenter, "Excitability and inhibitibility of motoneurons of different sizes," *Journal of neurophysiology,* vol. 28, no. 3, pp. 599-620, 1965.
[11] E. Henneman, G. Somjen, and D. O. Carpenter, "Functional significance of cell size in spinal motoneurons," *Journal of neurophysiology,* vol. 28, no. 3, pp. 560-580, 1965.
[12] J. L. Hicks, T. K. Uchida, A. Seth, A. Rajagopal, and S. L. Delp, "Is my model good enough? Best practices for verification and validation of musculoskeletal models and simulations of movement," *Journal of biomechanical engineering,* vol. 137, no. 2, p. 020905, 2015.
[13] N. Holt, J. Wakeling, and A. A. Biewener, "The effect of fast and slow motor unit activation on whole-muscle mechanical performance: the size principle may not pose a mechanical paradox," *Proceedings of the Royal Society B: Biological Sciences,* vol. 281, no. 1783, p. 20140002, 2014.
[14] B. Innocenti and F. Galbusera, *Human Orthopaedic Biomechanics: Fundamentals, Devices and Applications*. Academic Press, 2022.
[15] M. Jacquelin Perry, "Gait analysis: normal and pathological function," *New Jersey: SLACK,* 2010.
[16] M. Z. Kahraman and T. İŞLEN, "THE EFFECT OF MUSCLE FIBRIL TYPES ON PERFORMANCE IN FOOTBALL: A TRADITIONAL REVIEW," *International Journal of Education Technology and Scientific Researches,* vol. 8, no. 21, pp. 693-706, 2023.
[17] S. K. Kasereka *et al.*, "Equation-Based Modeling vs. Agent-Based Modeling with Applications to the Spread of COVID-19 Outbreak," *Mathematics,* vol. 11, no. 1, p. 253, 2023.
[18] K. M. Keyes, A. Shev, M. Tracy, and M. Cerdá, "Assessing the impact of alcohol taxation on rates of violent victimization in a large urban area: an agent-based modeling approach," *Addiction,* vol. 114, no. 2, pp. 236-247, 2019.
[19] D. V. Knudson and D. Knudson, *Fundamentals of biomechanics*. Springer, 2007.
[20] H. U. Kuriki, E. M. Mello, F. M. De Azevedo, L. S. O. Takahashi, N. Alves, and R. de Faria Negrão Filho, *The relationship between electromyography and muscle force*. Citeseer, 2012.
[21] S. Lee, M. Park, K. Lee, and J. Lee, "Scalable muscle-actuated human simulation and control," *ACM Transactions On Graphics (TOG),* vol. 38, no. 4, pp. 1-13, 2019.
[22] S.-H. Lee, E. Sifakis, and D. Terzopoulos, "Comprehensive biomechanical modeling and simulation of the upper body," *ACM Transactions on Graphics (TOG),* vol. 28, no. 4, pp. 1-17, 2009.
[23] M. Lempereur *et al.*, "A new deep learning-based method for the detection of gait events in children with gait disorders: Proof-of-concept and concurrent validity," *Journal of biomechanics,* vol. 98, p. 109490, 2020.
[24] W. Li, Q. Bai, and M. Zhang, "Agent-based influence propagation in social networks," in *2016 IEEE International Conference on Agents (ICA)*, 2016: IEEE, pp. 51-56.

[25] A. D. Likens and N. Stergiou, "Basic biomechanics," *Biomechanics and Gait Analysis,* p. 16, 2020.
[26] R. Lubaś, J. Wąs, and J. Porzycki, "Cellular Automata as the basis of effective and realistic agent-based models of crowd behavior," *The Journal of Supercomputing,* vol. 72, pp. 2170-2196, 2016.
[27] C. M. Macal and M. J. North, "Tutorial on agent-based modeling and simulation," in *Proceedings of the Winter Simulation Conference, 2005.*, 2005: IEEE, p. 14 pp.
[28] N. A. Mahoto, A. Shaikh, A. Sulaiman, M. S. Al Reshan, A. Rajab, and K. Rajab, "A machine learning based data modeling for medical diagnosis," *Biomedical Signal Processing and Control,* vol. 81, p. 104481, 2023.
[29] H. Milner-Brown, R. Stein, and R. Yemm, "The orderly recruitment of human motor units during voluntary isometric contractions," *The Journal of physiology,* vol. 230, no. 2, p. 359, 1973.
[30] A. Mirelman *et al.*, "Gait impairments in Parkinson's disease," *The Lancet Neurology,* vol. 18, no. 7, pp. 697-708, 2019.
[31] V. Modi, L. Fulton, A. Jacobson, S. Sueda, and D. I. Levin, "Emu: Efficient muscle simulation in deformation space," in *Computer Graphics Forum*, 2021, vol. 40, no. 1: Wiley Online Library, pp. 234-248.
[32] N. Montealegre and F. J. Rammig, "Agent-based modeling and simulation of artificial immune systems," in *2012 IEEE 15th International Symposium on Object/Component/Service-Oriented Real-Time Distributed Computing Workshops*, 2012: IEEE, pp. 212-219.
[33] F. R. Nezhat, M. Kavic, C. H. Nezhat, and C. Nezhat, "Forward We Go!," *JSLS: Journal of the Society of Laparoscopic & Robotic Surgeons,* vol. 27, no. 1, 2023.
[34] Y. Pang *et al.*, "Automatic detection and quantification of hand movements toward development of an objective assessment of tremor and bradykinesia in Parkinson's disease," *Journal of neuroscience methods,* vol. 333, p. 108576, 2020.
[35] E. J. Perreault, C. J. Heckman, and T. G. Sandercock, "Hill muscle model errors during movement are greatest within the physiologically relevant range of motor unit firing rates," *Journal of biomechanics,* vol. 36, no. 2, pp. 211-218, 2003.
[36] E. J. Perreault, T. G. Sandercock, and C. J. Heckman, "Hill muscle model performance during natural activation and electrical stimulation," in *2001 Conference Proceedings of the 23rd Annual International Conference of the IEEE Engineering in Medicine and Biology Society*, 2001, vol. 2: IEEE, pp. 1248-1251.
[37] B. P. Printy *et al.*, "Smartphone application for classification of motor impairment severity in Parkinson's disease," in *2014 36th Annual International Conference of the IEEE Engineering in Medicine and Biology Society*, 2014: IEEE, pp. 2686-2689.
[38] P. Rajpurkar, E. Chen, O. Banerjee, and E. J. Topol, "AI in health and medicine," *Nature medicine,* vol. 28, no. 1, pp. 31-38, 2022.
[39] M. Rana and M. Bhushan, "Machine learning and deep learning approach for medical image analysis: diagnosis to detection," *Multimedia Tools and Applications,* vol. 82, no. 17, pp. 26731-26769, 2023.
[40] M. Romeo, C. Monteagudo, and D. Sánchez-Quirós, "Muscle and fascia simulation with extended position based dynamics," in *Computer Graphics Forum*, 2020, vol. 39, no. 1: Wiley Online Library, pp. 134-146.
[41] N. A. Wani, R. Kumar, and J. Bedi, "DeepXplainer: An interpretable deep learning based approach for lung cancer detection using explainable artificial intelligence," Computer Methods and Programs in Biomedicine, vol. 243, p. 107879, 2024.
[42] H. Serhal, N. Abdallah, J.-M. Marion, P. Chauvet, M. Oueidat, and A. Humeau-Heurtier, "Overview on prediction, detection, and classification of atrial fibrillation using wavelets and AI on ECG," *Computers in Biology and Medicine,* vol. 142, p. 105168, 2022.
[43] H. E. C. d. Silva *et al.*, "The use of artificial intelligence tools in cancer detection compared to the traditional diagnostic imaging methods: An overview of the systematic reviews," *Plos one,* vol. 18, no. 10, p. e0292063, 2023.
[44] J. Slemenšek *et al.*, "Human gait activity recognition machine learning methods," *Sensors,* vol. 23, no. 2, p. 745, 2023.
[45] A. Solovyev *et al.*, "SPARK: a framework for multi-scale agent-based biomedical modeling," in *Proceedings of the 2010 spring simulation multiconference*, 2010, pp. 1-7.
[46] C. Steele, *Applications of EMG in clinical and sports medicine*. BoD–Books on Demand, 2012.
[47] S. J. Taylor, "Introducing agent-based modeling and simulation," in *Agent-based modeling and simulation*: Springer, 2014, pp. 1-10.
[48] M. Whittle, D. Levine, and J. Richards, *Whittle's gait analysis*. Butterworth-Heinemann, 2012.

[49]    D. A. Winter, *Biomechanics and motor control of human movement*. John wiley & sons, 2009.
[50]    L. Zhang, Z. Li, Y. Hu, C. Smith, E. M. G. Farewik, and R. Wang, "Ankle joint torque estimation using an EMG-driven neuromusculoskeletal model and an artificial neural network model," *IEEE Transactions on Automation Science and Engineering,* vol. 18, no. 2, pp. 564-573, 2020.